\documentclass[conference]{IEEEtran}
\ifCLASSINFOpdf
\else
\fi
\usepackage{amsmath,amssymb,amsfonts}
\usepackage{cite}
\usepackage{amsmath,amssymb,amsfonts}
\usepackage{algorithmic}
\usepackage{graphicx}
\usepackage{textcomp}
\usepackage{xcolor}
\usepackage{mathtools}
\usepackage{balance}
\usepackage{caption}
\usepackage{subcaption}
\usepackage{enumitem}
\usepackage{balance}

\DeclareMathOperator*{\argmin}{arg\,min}

\begin{document}

\title{Health Indicator Forecasting for Improving Remaining Useful Life Estimation}

\author{\IEEEauthorblockN{Qiyao Wang, Ahmed Farahat, Chetan Gupta, Haiyan Wang}
\IEEEauthorblockA{Industrial AI Laboratory, Hitachi America, Ltd. R$\&$D \\
Santa Clara, CA, USA \\
\{Qiyao.Wang, Ahmed.Farahat, Chetan.Gupta, Haiyan.Wang\}@hal.hitachi.com}}


%

\IEEEoverridecommandlockouts



\maketitle
\footnotetext[1]{978-1-7281-6286-7/20/\$31.00˜\copyright˜2020 IEEE. Personal use of this material is permitted.  Permission from IEEE must be obtained for all other uses, in any current or future media, including reprinting/republishing this material for advertising or promotional purposes, creating new collective works, for resale or redistribution to servers or lists, or reuse of any copyrighted component of this work in other works.}

\begin{abstract}
Prognostics is concerned with predicting the future health of the equipment and any potential failures. With the advances in the Internet of Things (IoT), data-driven approaches for prognostics that leverage the power of machine learning models are gaining popularity. One of the most important categories of data-driven approaches relies on a predefined or learned health indicator to characterize the equipment condition up to the present time and make inference on how it is likely to evolve in the future. In these approaches, health indicator forecasting that constructs the health indicator curve over the lifespan using partially observed measurements (i.e., health indicator values within an initial period) plays a key role. Existing health indicator forecasting algorithms, such as the functional Empirical Bayesian approach, the regression-based formulation, a naive scenario matching based on the nearest neighbor, have certain limitations. In this paper, we propose a new `generative + scenario matching' algorithm for health indicator forecasting. The key idea behind the proposed approach is to first non-parametrically fit the underlying health indicator curve with a continuous Gaussian Process using a sample of run-to-failure health indicator curves. The proposed approach then generates a rich set of random curves from the learned distribution, attempting to obtain all possible variations of the target health condition evolution process over the system's lifespan. The health indicator extrapolation for a piece of functioning equipment is inferred as the generated curve that has the highest matching level within the observed period. Our experimental results show the superiority of our algorithm over the other state-of-the-art methods.

\end{abstract}


%
\IEEEpeerreviewmaketitle

\section{Introduction}
\label{sec1}

Prognostics is concerned with predicting the future health of the equipment and any potential failures. Prognostics techniques are typically applied when a fault or degradation is detected to predict when a failure or severe degradation is going to happen. Based on the type of information used, prognostics approaches can be categorized into model-based, data-driven and hybrid which can be described as follows:

\begin{itemize}[leftmargin=*]
\item \textbf{Model-based approaches} use a physical model to represent and simulate the degradation of a dynamical system until the failure point to estimate the time-to-failure. Model-based prognostics can be easily explained to a domain expert to justify a prediction and they need relatively fewer failure data to be trained. On the other hand, it is difficult to model degradation in complex systems. These models are also limited to sensor data so other data types such as free text or images cannot be incorporated.
\item \textbf{Data-driven approaches} use historical system measurements of run-to-failure examples to estimate the time-to-failure or the probability of failures for new equipment. In comparison to model-based prognostics, data-driven techniques are equipment-agnostic which makes them faster to build and deploy. They are also capable of incorporating heterogeneous types of data such as events data. However, these models need much more data and their outputs are more difficult to interpret by domain experts.
\item \textbf{Hybrid approaches} use both physics-based models and historical system measurements. For instance, learning the parameter of a physical model from the data or using physical-based features as input to a data-driven method.
\end{itemize}

The focus of this paper is on data-driven approaches for prognostics. Data-driven prognostics can be achieved by either (i) learning a direct mapping from the raw sensor measurements to time-to-failure estimates, or (ii) extrapolate a health indicator that reflects the equipment degradation to predict when the equipment is going to reach a certain health state. The first approach is usually more accurate when a large number of historical failures are available as it does not impose a constraint on the complexity of the learned decision rules but this makes these rules very difficult to explain to domain experts. On the other, the second approach is more desirable for domain experts as it gives insights into the health of the equipment through indicators that they can understand and relate to which makes the decision making much easier. This paper is concerned with health indicator-based prognostics.

Health indicators are typically associated with how close the equipment to its end-of-life and they can be either (i) defined based on domain-knowledge or (ii) learned from the data. Domain-based health indicators are usually defined using one or more of the degradation signals (e.g., temperature) that are measured during equipment operation, or computed from one or more of the raw sensor measurements (e.g., the cooling capacity of a chiller). On the other hand, many methods have been proposed to learn health indicators from data. These methods include signal processing-based methods \cite{lei2018machinery}, fusion of hand-crafted indicators \cite{liu2013data}, using Self Organizing Maps (SOM) \cite{qiu2003robust}, using Hidden Markov Models (HMM)s \cite{ramasso2009contribution}, and recently using deep reinforcement learning \cite{zhangequipment}.


After a health indicator is defined or learned, it needs to be extrapolated over the lifespan to predict when the failure is going to happen. The achieved health indicator curve characterizes the status of the equipment up to the present and provides information on how the equipment health is likely to evolve in the future. Two sources of information are often provided to conduct health curve forecasting for a functioning equipment, including the target equipment partially realized health indicator data and the run-to-failure health indicator curves (i.e., from the beginning of life to a complete failure state) for a population of equipment \cite{zhou2011degradation, zhangequipment}.

There is a significant amount of research on health indicator forecasting. Under the assumption that only the health information of the considered equipment is accessible, methods including the time series forecasting-based approach \cite{rehfeldt1987evaluation} and the exponential smoothing technique \cite{Lu2001real} have been investigated. These approaches are known to have low accuracy as they fail to incorporate useful knowledge regarding the overall health indicator trend in the run-to-failure data. 



A review of the relevant literature that combines the two sources of information is provided as follows. The health indicator forecasting task is popularly formulated from the Empirical Bayesian perspective \cite{gebraeel2006sensory, gebraeel2005residual, zhou2011degradation}. It is often assumed that the underlying health indicator curve over time follows an unknown continuous stochastic process. These approaches use the run-to-failure data samples to fit the statistical distribution, which is treated as prior knowledge. Next, for new equipment, they update the knowledge by calculating the posterior health indicator distribution conditional on the target equipment initial health data. The achieved posterior distribution is then used to infer the considered equipment health indicator evolution curve over its lifespan. Most of the traditional work in this area assume certain parametric forms for the mean and covariance function of the underlying stochastic process (e.g., linear or exponential mean trend, and compound covariance), which makes the methods applicable to limited scenarios \cite{gebraeel2006sensory, gebraeel2005residual}. A non-parametric health evolution model is later considered \cite{zhou2011degradation}. However, the calculated posterior distribution for the new equipment is valid only when the random errors between the raw measurements of the health curve and the projection scores of the functional principal component analysis (PCA) (which is a counterpart of the traditional PCA for continuous stochastic processes [11], [12]) are jointly distributed as multivariate Gaussian. Theoretically, the constructed health indicator curve will be biased when the actual data does not demonstrate this required property. This phenomenon was observed in our numerical experiments. 

Another promising approach is to formulate the health indicator forecasting task as a regression problem. Specifically, by building a machine learning model that outputs the health indicator at the next one or more time points using a sequence of past values as the input, one can fully extrapolate the new equipment health curve from the partially realized data \cite{ho2002comparative, qiu2014ensemble}. This formulation fails to model the complete health indicator evolution pattern over the lifespan, as it often pre-processes each individual health indicator curve by cutting it into a set of small windows to extract the required training data. Furthermore, the processed data are inappropriately treated as independent samples even though they are extracted from the different time periods of the same equipment.


To overcome the aforementioned challenges, in the paper, we propose a new framework for health indicator forecasting. We propose to first non-parametrically estimate the mean and covariance patterns of the underlying stochastic health indicator curve using the run-to-failure data samples. This is the prior distribution learning step considered in \cite{zhou2011degradation}. Instead of attempting to analytically compute the posterior distribution for new equipment, we propose a model-free `posterior' updating strategy. Under the assumption that the health indicator follows a Gaussian Process with the learned mean and covariance, which is less restrictive requirement than \cite{zhou2011degradation}, we generate random curves from the learned distribution to cover all possible scenarios and, for each new equipment, identify the final forecasting as the generated health indicator curve with the highest predefined matching score for the observed period. The proposed `generative + scenario matching' method is less restrictive than the Empirical Bayesian approaches \cite{gebraeel2006sensory, gebraeel2005residual, zhou2011degradation} and is more efficient than the regression-based formulation. The superior performance of our proposal is demonstrated by the numerical experiments in Section \ref{sec4}. 


The rest of the paper is organized as follows. Section \ref{sec2} presents the problem definition, and the proposed `generative + scenario matching' method. Section \ref{sec3} provides the inference of remaining useful life from the forecasted health indicator curve. Section \ref{sec4} describes our experiment on a benchmark data set. Section \ref{sec5} concludes the paper.

\section{Proposed Method for Health Indicator Forecasting}
\label{sec2}
\subsection{Notations and Problem Definition}
\label{sec2.1}
Suppose that we have access to the health indicator data of $N$ equipment of the same type. For the $i$-th equipment, $i=1,...,N$, the observed health indicator curve is denoted as $\mathbf{S}_i=[S_i(t_{i,1}),...,S_i(t_{i,m_i})]^T$, where $m_i$ is the number of observations and $[t_{i,1},...,t_{i,m_i}]^T$ are the corresponding observation time within a bounded time domain $[0,M]$. Note that $M<\infty$ represents the longest possible life time of the considered type of equipment. 

For each of the equipment, the observed health indicator curve can be both complete and incomplete. A complete health indicator curve is a continuously observed signal from the beginning of life to a failure state \cite{zhou2011degradation}. Under this scenario, the number of observation per equipment $m_i$ is relatively large and $[t_{i,1},...,t_{i,m_i}]^T$ are regularly scattered within $[0,M]$.  An incomplete health indicator curve is a signal that consists of intermittently observed health condition data or continuously observed signals within short intervals over the life cycle \cite{zhou2011degradation, wang2019multilayer, wang2017two}. This means that $[t_{i,1},...,t_{i,m_i}]^T$ are sparsely distributed over $[0,M]$ or clustered within narrow intervals within $[0,M]$. Note that even though the health indicator curve of individual equipment can be incomplete, it is required that the combined signal across all equipment is complete \cite{wang2019multilayer}, i.e., $[t_{1,1},...,t_{1,m_1}, ..., t_{N,1},...,t_{1,m_N}]^T$ are densely and regularly scattered within $[0,M]$.

Suppose that we are also provided the health indicator measurements of functioning equipment within the initial period of its life. Mathematically, the obtained health indicator signal is denoted as $\mathbf{S}_{new}=[S_{new}(t_{new, 1}), ..., S_{new}(t_{new, m_{new}})]^T$, where each element of $[t_{new, 1},...,t_{new, m_{new}}]^T$ falls within $[0, M^*]$, $0 < M^*< M$. 

The problem of interest is to forecast the health indicator signal for the operating equipment over its lifespan (i.e., $[0, M]$) based on the provided two sources of information: the historical health indicator curves over $[0, M]$, and the functioning equipment initial measurements $\mathbf{S}_{new}$ within $[0, M^*]$.

\subsection{Outline of the Proposed Method}
\label{sec2.2}
The proposed approach is based on the idea presented by \cite{zhou2011degradation} that the health indicator curve of any equipment is a random sample from an underlying stochastic process within the considered time range $[0, M]$. Suppose that we can generate a large number of health indicator signals with support $[0, M]$ from the underlying distribution, there is a high probability that at least one of the generated health indicator curves is close to that of the new equipment. Under the intuition that curves that are more consistent with the actual health indicator within $[0, M^*]$ are more likely to be close to the ground truth over the lifespan, we propose to calculate the matching level of each candidate with the actual values within the observed period and outputs the one with the highest matching score as the final forecasting. 

\begin{figure}[htbp]
\centering
\includegraphics[width=74mm]{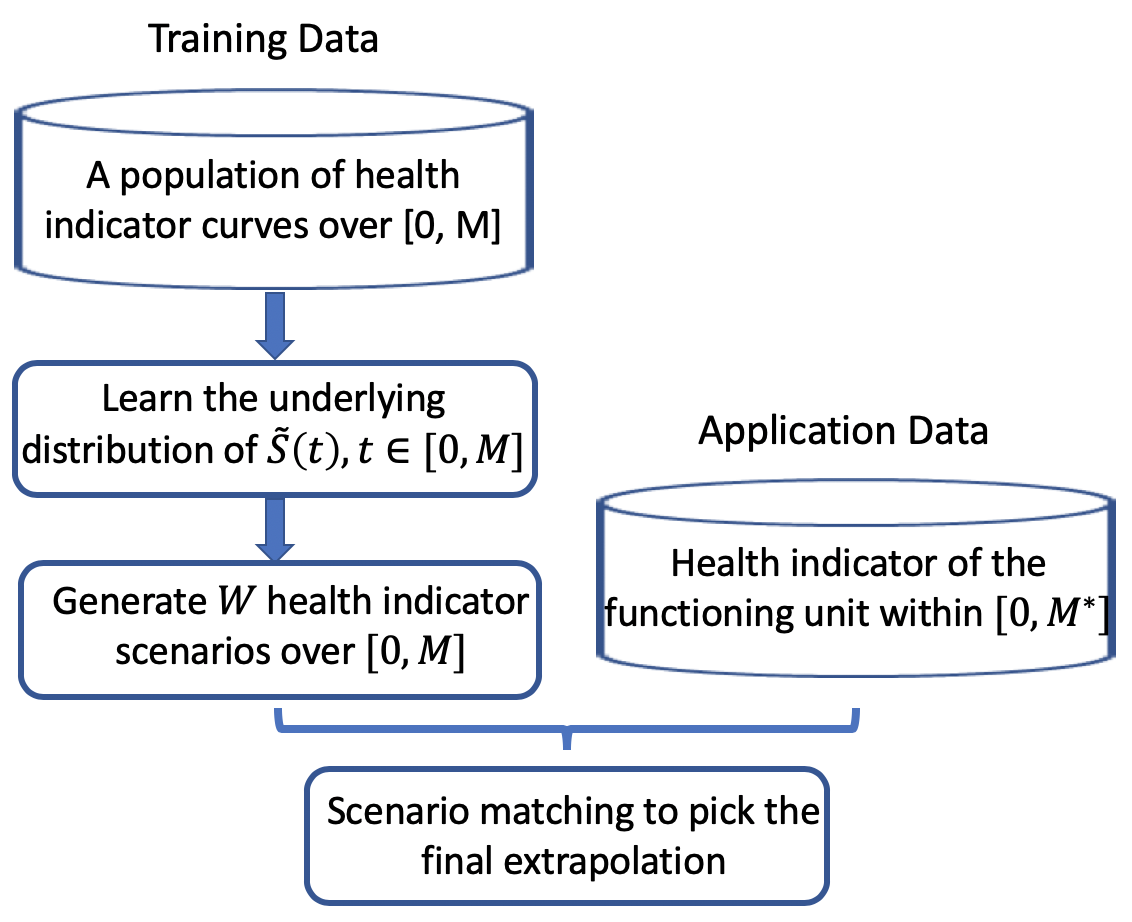}
\caption{Flow chart for the proposed method.}
\label{Fig:flow_chart}
\vspace{-0.1in}
\end{figure}

The proposed `generative + scenario matching' approach is outlined by the flow chart in Fig.~\ref{Fig:flow_chart}. There are two major components in the proposed algorithm. The generative modeling step that learns the data distribution and generates realizations of health indicator curves over the support is presented in Section \ref{sec2.3}. The scenario matching module that selects the final forecasting among all the simulated candidates is described in Section \ref{sec2.4}. 

\subsection{Gaussian Process-Based Generative Model}
\label{sec2.3}
Following the formulation in the prior art \cite{gebraeel2006sensory, gebraeel2005residual, zhou2011degradation}, for each equipment in the training set, the observed run-to-failure health data are assumed to be discrete realizations of an underlying continuous random processes that are contaminated by random errors. Specifically, for the $i$-th equipment, let the underlying health indicator curve be $\tilde{S}_i(t)$ and the zero-mean random error curve be $\epsilon_i(t)$ for any $t\in[0,M]$. The observed health indicator values $\mathbf{S}_i$ are then finite realizations of the signal in Eq.~(\ref{decomposition}) at time points $[t_{i,1},...,t_{i,m_i}]^T$.
\begin{equation}\label{decomposition}
S_i(t) = \tilde{S}_i(t) + \epsilon_i(t), t\in[0,M]
\end{equation}
Note that the underlying health indicator curves $\{\tilde{S}_i(t)\}_{i=1}^N$ can be viewed as i.i.d random samples from a continuous stochastic process with an unknown mean function $\mu(t)$ (i.e., the overall underlying health indicator trend) and an unknown covariance function $G(t, t^\prime)$, $t, t^\prime \in [0,M]$ (i.e., the random deviation from the overall health indicator trend). 

To recover the underlying stochastic distribution, we need to estimate $\mu(t)$ and $G(t, t^\prime)$ using the observed data $\{\mathbf{S}_i=[S_i(t_{i,1}),...,S_i(t_{i,m_i})]^T\}_{i=1}^N$. Following the strategy in \cite{zhou2011degradation}, we propose to non-parametrically construct $\mu(t)$ and $G(t, t^\prime)$ without constraining our considerations to any parametric shapes. The generalized data fitting approach encompasses the special cases that assume parametric mean and covariance functions, such as the linear trend $\mu(t)=\alpha+\beta t$ where $\alpha\sim N(0, \delta_{\alpha})$ and $\beta\sim N(0, \delta_{\beta})$ and the covariance function with the formula $G(t, t^\prime)=\delta_{\alpha} + \delta_{\beta}tt^\prime$ \cite{gebraeel2006sensory, gebraeel2005residual}.


According to functional data analysis \cite{ramsay2006functional, yao2005functional}, depending on the characteristic of the historical health indicator measurements, there are two ways of estimating the mean and covariance functions. When all the $N$ health indicator signals are complete signals, i.e., $[t_{i,1},...,t_{i,m_i}]^T$ are densely and regularly scattered within $[0,M]$ for all $i=1,...,N$, non-parametric interpolation techniques such as local linear or quadratic smoothing are first adopted to obtain a consistent recovery for the underlying continuous curve $\tilde{S}_i(t)$ using $\mathbf{S}_i$ \cite{ramsay2006functional}. Let's denote the resulting recovery as $\hat{\tilde{S}}_i(t)$. The mean and covariance functions are estimated by their sample counterparts presented in Eq.~(\ref{est1}). 
\begin{equation} \label{est1}
\begin{split}
\centering
\hat{\mu}_c(t)& = \frac{1}{N}\sum_{i=1}^N \hat{\tilde{S}}_i(t)\\
\hat{G}_c(t, t^\prime) & = \frac{1}{N-1}\sum_{i=1}^N[\hat{\tilde{S}}_i(t)-\hat{\mu}_c(t)][\hat{\tilde{S}}_i(t^\prime)-\hat{\mu}_c(t^\prime)]
\end{split}
\end{equation}

When not all the $N$ health indicator signals are complete, the individual curves can no longer be consistently recovered by the limited amount of information in the raw data $\mathbf{S}_i$. Under the assumption that the combined signal across all equipment is complete, the mean and covariance function can be estimated by borrowing information across all the equipment \cite{yao2005functional, zhou2011degradation}, i.e., using the pooled sample $[\mathbf{S}_1,..., \mathbf{S}_N]^T$. The local linear smoothing technique have been theoretically shown to be consistent under mild regularization assumptions. To be more specific, for any target time $t\in[0, M]$, we define the local linear smoother of $\mu(t)$ by minimizing
\begin{equation}\label{ls1}
\sum_{i=1}^{N}\sum_{j=1}^{m_{i}}K_1(\frac{t_{i,j}-t}{h_{\mu}})[S_i(t_{i,j})-\beta_0-\beta_{1}(t-t_{i,j})]^2
\end{equation}
with respect to $\beta_0$ and $\beta_1$, where $K_1(\cdot)$ is a smoothing kernel and $h_{\mu}$ is the bandwidth, yielding $\hat{\mu}_{ic}(t)=\hat{\beta}_0(t)$. Similarly, for any target $(t, t^\prime)\in [0,M]\times [0,M]$, we define the local linear smoother of $G(t, t^\prime)$ by minimizing
\begin{multline}
    \label{ls2}
   \sum_{i=1}^{N}\sum_{1 \leq j\neq l \leq m_i}^{m_{i}}K_2(\frac{t_{i,j}-t}{h_{G}}, \frac{t_{i,l}-t^\prime}{h_{G}})[G(t_{i,j}, t_{i,l})\\ 
    -(\gamma_0 + \gamma_{11}(t-t_{i,j}) + \gamma_{12}(t^\prime-t_{i,l}))]^2
 \end{multline}
 with respect to $\gamma_0$, $\gamma_{11}$ and $\gamma_{12}$, where $K_2(\cdot, \cdot)$ is a two-dimensional smoothing kernel and $h_G$ is the bandwidth, yielding $\hat{G}_{ic}(t, t^\prime)=\hat{\gamma}_0(t, t^\prime)$. Note that the estimators in Eq.~(\ref{ls1}),  (\ref{ls2}) is equivalent to those in Eq.~(\ref{est1}) when all the health indicator signals are complete. This indicates that the estimators in Eq.~(\ref{ls1}),  (\ref{ls2}) are applicable for both complete and incomplete data scenarios. In the paper, we use this approach to obtain the non-parametric estimates for the mean and covariance functions.

As the next step, we aim to generate a set of random health indicator curves from the stochastic process with estimated non-parametric mean and covariance functions. To obtain the close form for the data generation, we need to introduce the functional principal component analysis  \cite{hall2006properties, yao2005functional}. For the covariance function $G(t, t^\prime)$, the eigenvalues $\{\lambda_r\}_{r=1}^\infty$ and eigenfunctions $\{\phi_r(t)\}_{r=1}^\infty$ are solutions of equation
\begin{equation}\label{fpac1}
\lambda \phi(t) = \int_{t^\prime} G(t, t^\prime)\phi(t^\prime) dt^\prime.
\end{equation}
Note that $\{\phi_r(t)\}_{r=1}^\infty$ is an orthonormal basis on $L^2([0,M])$, i.e., 
\begin{equation}\label{fpac2}
\begin{split}
\int_{t} \phi_r^2(t) dt & =1, r=1,...,\infty \\
\int_{t} \phi_r(t) \phi_{r^\prime}(t)dt & =0, 1\leq r \neq r^\prime \leq \infty.
\end{split}
\end{equation}
Different approaches have been proposed to solve the Eq.~(\ref{fpac1}) \cite{yao2005functional, zhou2011degradation}. By plugging the eigenfunctions into the Karhunen-Lo\`eve expansion  \cite{ramsay2006functional}, we know that a random process with mean $\mu(t)$ and covariance $G(t, t^\prime)$ can be represented as 
\begin{equation}\label{fpac3}
\tilde{S}(t)=\mu(t) + \sum_{r=1}^{\infty}[\int_{t}(\tilde{S}(t)-\mu(t))\phi_r(t)dt] \phi_r(t), 
\end{equation}
where $\int_{t}(\tilde{S}(t)-\mu(t))\phi_r(t)dt$ is a random variable with mean 0 and variance $\lambda_r$, the $r$-th largest eigenvalue of $G(t, t^\prime)$. Typically, the underlying health indicator curve $\tilde{S}(t)$ is relatively smooth and Eq.~(\ref{fpac3}) is well approximated by a truncated version, 
\begin{equation}\label{fpac4}
\tilde{S}(t)\approx \mu(t) + \sum_{r=1}^{P}\xi_{r}\phi_r(t), 
\end{equation}
where $\xi_{r}=\int_{t}(\tilde{S}(t)-\mu(t))\phi_r(t)dt$. Note that $P$ can be determined by the cross-validation approach, the percentage of variance explained method, and other penalty criteria such as AIC and BIC \cite{yao2005functional}. 

When $\tilde{S}(t)$ is a Gaussian Process, $\xi_{r}=\int_{t}(\tilde{S}(t)-\mu(t))\phi_r(t)dt$ is known to follow $N(0, \lambda_r)$. Under the Gaussian Process assumption, we propose generate $W$ health indicator curves over $[0,M]$ using Eq.~(\ref{fpac4}) as follows. We first conduct functional principal component analysis on $\hat{G}(t,t^\prime)$ to obtain $\{\hat{\lambda}_r, \hat{\phi}_r(t)\}_{r=1}^P$ \cite{hall2006properties, yao2005functional}. Then for $i=1,...,W$, 
\begin{enumerate}[leftmargin=*]
\item Draw random numbers from $N(0, \hat{\lambda}_r)$ for $r=1,...,P$, denoted as $\xi_{i,1},...,\xi_{i,P}$.
\item Plug $\xi_{i,1},...,\xi_{i,P}$, $\hat{\mu}(t)$, and  $\{ \hat{\phi}_r(t)\}_{r=1}^P$ into Eq.~(\ref{fpac4}) to obtain $\tilde{S}_i(t)$.
\end{enumerate}
Note that $\{\xi_{i,1},...,\xi_{i,P}\}$ represent the random deviation of each realization $\tilde{S}_i(t)$ from the common population level characteristics $\hat{\mu}(t)$, and  $\{ \hat{\phi}_r(t)\}_{r=1}^P$.


\subsection{Health Indicator Forecasting by Scenario Matching}
\label{sec2.4}
In this section, we propose a systematic way of identifying the final forecasting among all the $W$ simulated random samples in Section \ref{sec2.3}. We propose to quantify the matching level for each simulated health indicator curve by the root mean squared error with the new equipment actual observations within $[0,M^*]$, where $0 < M^*< M$. Mathematically, for $i=1,...,W$, the matching level $H_{i, M^*} \in \mathbb{R}^+$ is calculated by 
\begin{equation}\label{match1}
H_{i, M^*} = \sqrt{\frac{1}{m_{new}}\sum_{j=1}^{m_{new}}(\tilde{S}_i(t_{new, j})-S_{new}(t_{new, j}))^2}, 
\end{equation}
where $S_{new}(t_{new, j})$ is the actual health data of the operating equipment at time $t_{new, j}\in[0,M^*]$, and $\tilde{S}_i(t_{new, j})$ is the evaluated value of the $i$-th simulated curve at the same time. 

Given the calculated matching scores, we pick the random curve with the highest matching level, i.e., the lowest $H_{i, M^*}$, as the forecasted health indicator for the target equipment. Mathematically, the final forecasting $\hat{\tilde{S}}_{new}(t)$ is given by 
\begin{equation}\label{match1}
\hat{\tilde{S}}_{new}(t) = \tilde{S}_{sel}(t), \text{where   } sel=\argmin_{i}H_{i, M*}.  
\end{equation}

\subsection{Relationship to Prior Work}
\label{sec2.5}
The idea of extrapolating a partially observed time series data by matching it with a set of fully observed time series has been shown to be powerful in several applications  \cite{sridevi2018effective, zhou2009scenario}. In the prior art, the candidate scenarios are usually not constructed from statistical distribution perspectives. For instance, a set of possible scenarios is obtained by fitting the data with different configurations of ARIMA \cite{sridevi2018effective}. Under such circumstances, there is no guarantee that a good match exists among the candidate set. In our proposal, we learn the statistical distribution of the underlying process from which the data for all equipment are generated. We propose to generate candidate scenarios from the learned distribution, which better ensures the quality of the forecasting result. 

Compared to the method in \cite{zhou2011degradation} that attempts to get a closed form for the posterior distribution of the health indicator curve, the proposed model-free approach is more general and is applicable to a wider range of problems. The relatively strict requirement on the joint distribution of measurement errors in $\mathbf{S}_{new}$ and the FPCA projection scores $\{\int \tilde{S}_{new}(t)\phi_r(t)dt\}_{r=1}^P$ makes the approach \cite{zhou2011degradation} tend to produce biased extrapolating results in a lot of real applications.


\section{Remaining Useful Life Prediction}
\label{sec3}
In this section, we consider deploying the achieved health indicator curve in Section \ref{sec2} to predict the remaining time to the occurrence of a soft-failure, which is referred to as the remaining useful life (RUL) prediction problem in this paper. Following the prior art \cite{gebraeel2006sensory, gebraeel2005residual, zhou2011degradation}, we define soft-failure as the failure that occurs when the health indicator reaches a pre-specified critical threshold. Predicting the soft-failures gives the maintenance crew sufficient time to take appropriate actions such as repairs and component replacement before more severe failures happen. 

Without loss of generality, let's assume that higher health indicator values corresponding to better health conditions. Similar to most of the prior art in the health indicator area \cite{gebraeel2006sensory, gebraeel2005residual, zhou2011degradation}, it is assumed that the underlying health indicator curve is monotonic. Let's denote the underlying health indicator curve of the operating equipment by $\tilde{S}_{new}(t)$, then the true soft-failure time $T$ is
\begin{equation}\label{RUL1}
T = \inf_{t\in[0,M]} \{\tilde{S}_{new}(t) \leq \theta\}, 
\end{equation}
where $\theta$ is the pre-determined failure threshold. In this paper, we assume that $\theta$ is given and its value is reasonably set to ensure that the failure time $T$ exists, which means that $\theta$ should not be too small such that $\tilde{S}_{new}(t) \leq \theta$ will not be satisfied for any $t\in [0, M]$. In real applications, domain experts can typically provide good threshold values based on their subjective judgment and well-accepted standards in the domain. 

The last health indicator evaluation time for a new equipment is $t_{new, m_{new}}$, which will be denoted as $t^*$ in the later discussions. It is assumed that $T>t^*$, i.e., the new equipment has not encountered any soft-failures up to the last observation time. This is a reasonable assumption as prediction for RUL is not need if the soft-failure has happened within the observed period $[0,t^*]$. The ground truth value of RUL at time $t^*$ is 
\begin{equation}\label{RUL2}
\text{RUL}_{new, t^*} = \inf_{t\in[t^*,M]} \{\tilde{S}_{new}(t) \leq \theta\} - t^*.
\end{equation}
Based on the forecasted health indicator curve $\hat{\tilde{S}}_{new}(t)$ in Section \ref{sec2}, we propose the following point estimator for $\text{RUL}_{new, t^*} $ 
\begin{equation}\label{RUL3}
\hat{\text{RUL}}_{new, t^*} = \inf_{t\in[t^*,M]} \{\hat{\tilde{S}}_{new}(t) \leq \theta\} - t^*.
\end{equation}

The performance of our proposed RUL estimator in Eq.~(\ref{RUL3}) is demonstrated by numerical experiments on a benchmark data set in the next section.

\section{Experiments on C-MAPSS Data Set}
\label{sec4}
In this section, we apply the proposed health indicator forecasting approach to a widely-used benchmark data set called NASA C-MAPSS (Commercial Modular Aero-Propulsion System Simulation) data \cite{saxena2008phm08}, in comparison with three alternative state-of-the-art approaches (see Section \ref{sec4.3}). For each method, the forecasted health indicator curve is utilized to produce the corresponding RUL estimations. As shown by the experimental results, the proposed `generative + scenario matching' approach significantly outperforms all these alternative methods in terms of both health indicator forecasting and RUL estimation.  

\subsection{Background}
\label{sec4.1}
C-MAPSS data set has been popularly used to justify performances for remaining useful life estimation tasks \cite{ zheng2017long, wang2019remaining}. It contains 21 simulated sensor signals, 3 operating setting variables for a group of turbofan engines as they run until critical failures. There are four data subsets in C-MAPSS that correspond to scenarios with different numbers of operating conditions and fault modes \cite{saxena2008phm08}. Each subset is divided into the training and testing sets. The training sets contain run-to-failure data for a set of engines that have been continuously monitored from an initial healthy state to a failure state. The testing sets consist of prior-to-failure data where all the data are truncated at a certain time before failure. Tab.~\ref{bg} provides a summary for each subset in C-MAPSS. In the experiment, we only consider the first two subsets where there is only one fault mode and the health indicator curves can be assumed to come from the same distribution, as most health indicator analytical methods implicitly or explicitly require that the health data curves are homogeneous. On way to apply the proposed method to the last two subsets with two failure modes is to cluster the health indicator curves into two groups, each with a monotonic trend, then apply the proposed method on each cluster. This is the subject of a future work.


\begin{table}[htbp]
\vspace{-0.08in}
\caption{Summary of the subsets in C-MAPSS data set.}
\vspace{-0.15in}
\begin{center}
\begin{tabular}{c|cccc}
\hline
\hline
\textbf{}& \textbf{FD001}&   \textbf{FD002}&  \textbf{FD003}& \textbf{FD004}\\
\hline
$\#$ of engines in training& 100 & 260 & 100  & 249  \\
$\#$ of engines in testing& 100 & 259 & 100 &  248 \\
$\#$ of operating conditions& 1 & 6 & 1 &   6 \\
$\#$ of fault modes & 1 & 1 & 2 &  2 \\
\hline
\hline
\end{tabular}
\vspace{-0.17in}
\label{bg}
\end{center}
\end{table}

\subsection{Data Preparation}
\label{sec4.2}
\begin{figure}[htbp]
	\centering
	\begin{subfigure}[t]{1.625in}
		\centering
        \caption{Degradation signal}\label{good}
        \vspace{-0.1in}
		\includegraphics[width=36mm]{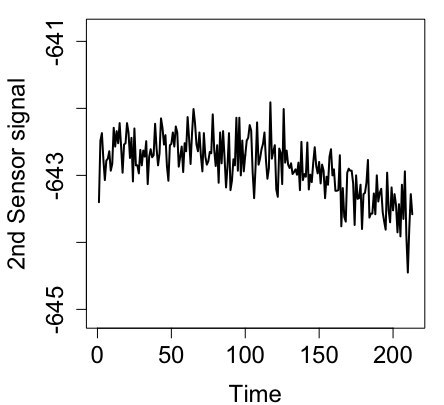} 		
	\end{subfigure}
	\quad
	\begin{subfigure}[t]{1.625in}
		\centering
        \caption{Irrelevant sensor}\label{bad}
        \vspace{-0.1in}
		\includegraphics[width=36mm,]{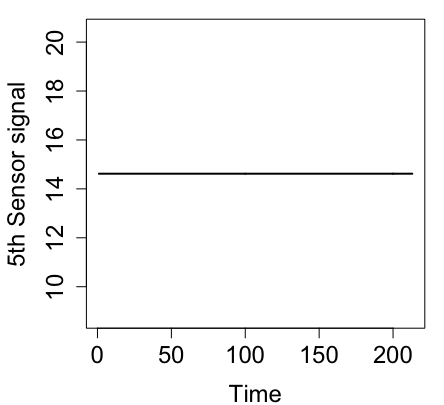} 		
	\end{subfigure}
	\vspace{-0.1in}
	\caption{Examples of a degradation signal and a sensor signal that is irrelevant to the degradation for an engine in FD001. }\label{exp}
\end{figure}

\textit{1) Health indicator definition:} In prognostic, the sensor signals that are evolve in a manner that is related to the degradation process are known as degradation signals \cite{liu2013data, gebraeel2006sensory, gebraeel2005residual}. Examples of a degradation signal and a sensor signal that is not indicative of the degradation are given in Fig.~\ref{exp}. Practically, sensor signals whose values at the end of life significantly deviate from those at the initial life period are identified as degradation signals \cite{liu2013data}. Based on this rule, we can identify a set of degradation signals in the C-MAPSS data. Each of these degradation signals is treated as a domain knowledge-based health indicator in the experiment. Without loss of generality, we multiply the raw sensor data by $-1$ when the sensor exhibits an increasing trend to make all considered health indicators decrease over time. 

\textit{2) Training and testing splitting:} To evaluate the performance of health indicator forecasting models, we need to know the actual health indicator over the equipment lifespan, which is not available for the truncated parts of the original C-MAPSS testing data. Consequently, we propose to split the run-to-failure data in the training sets of C-MAPSS into new training and testing sets. Engineering systems often decay at various rates and have different lengths of life. For instance, in FD001, the lifetime ranges from 128 to 362 across the 100 engines. To balance the lifetime distribution between the training and testing, we propose a stratified random data splitting strategy. For a given data set, we group all the engines into five clusters according to the lengths of their lifespans: the engines with a lifespan shorter than the lower $20\%$ percentile of all lifetimes form the first group; the engines with a lifespan between the lower $20\%$ and $40\%$ percentiles make the second group; the remaining three clusters can be created analogously. Next, we use the simple random sampling to assign $70\%$ engines within each cluster to training and the remaining $30\%$ engines into testing. 

\begin{figure}[htbp]
\centering
\includegraphics[width=38mm]{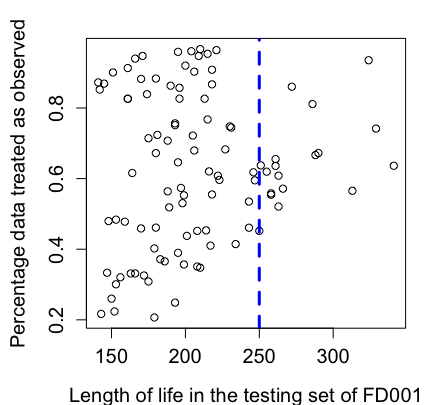}
\vspace{-0.05in}
\caption{Investigation on test engine's data truncation strategy in the orignal C-MAPSS data set.}
\label{Fig:ratio}
\end{figure}

\textit{3) Stratified random truncation for engines in testing:} For each engine assigned to the testing sets, we need to divide its lifespan into an observed and unobserved period, i.e., determine how much data should be treated as given information for the health indicator forecasting problem. We propose to follow the data truncation strategy in the original C-MAPSS testing set. In Fig.~\ref{Fig:ratio}, the x-axis represents the actual length of life and the y-axis is the percentage of data given in the original testing set of FD001. It can be seen that the percentage of data supplied is approximately uniformly distributed between $20\%$ and $97\%$ when the actual length of life is shorter (i.e., on the left side of the blue vertical line), while the percentage of the observed data is much higher (between $55\%$ and $97\%$) when the engine survives longer (i.e., on the right side of the blue vertical line). Based on this observation, we propose conduct a stratified random data truncation. For a considered testing set, we check the actual lifetime of each engine and identify the lower $80\%$ percentile. Next, for the $i$-th engine in the testing set, if its length of life is smaller than the $80\%$ percentile, we randomly generate $r_i$ from Uniform$[0.2, 0.97]$, otherwise from Uniform$[0.6, 0.97]$. The first $100r_i\%$ of the $i$-th equipment health indicator measurements are then treated as observed when conducting the health indicator forecasting.  

\textit{4) Removing the effect of operating conditions:}
According to Tab.~\ref{bg}, there are six operating conditions represented by the three numerical operating condition variables in FD002. The inconsistency in operating conditions across the lifespan makes it inappropriate to directly use the raw sensor data as the degradation signal/health indicator. 

\begin{figure}[htbp]
	\centering
	\begin{subfigure}[t]{1.65in}
		\centering
        \caption{Raw signal for a randomly selected engine in FD002.}\label{raw}
        \vspace{-0.08in}
		\includegraphics[width=36.5mm, height=32mm]{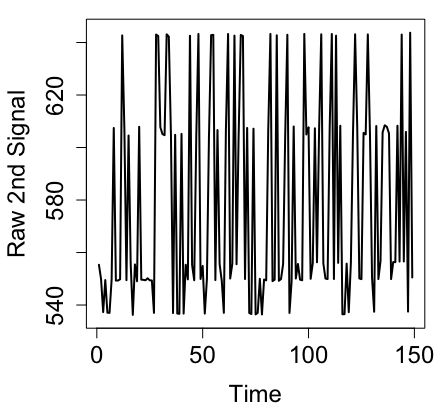}		
	\end{subfigure}
        \quad
	\begin{subfigure}[t]{1.65in}
		\centering
        \caption{Normalized signal after removing operating conditions.}\label{normalized}
        \vspace{-0.08in}
		\includegraphics[width=36.5mm, height=32mm]{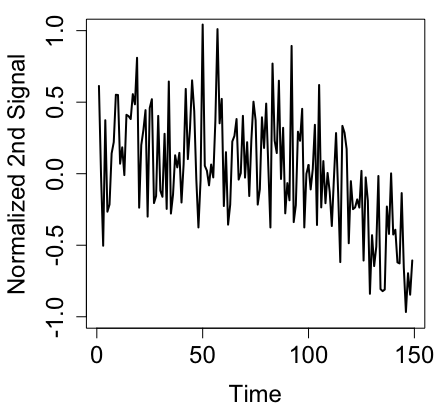}		
	\end{subfigure}
	\vspace{-0.05in}
	\caption{Removing the effect of operating conditions on data. }\label{normalization}
	\vspace{-0.15in}
\end{figure}

To remove the impact of operating conditions on the health indicator, we follow the regression-based data normalization strategy proposed by \cite{wang2019remaining, wang2018maintenance, wang2019evaluation}. For each of the selected degradation signal, we use the data from each training set to train a regression model which maps from the operating condition variables to the sensor variable. The achieved regression model enables us to estimate the would-be sensor data given any operating condition. We then calculate the normalized sensor data by deducting the would-be sensor data from the raw sensor readings. In our experiment, we train the regression model using Multilayer Perceptron. An example of the raw sensor data and the normalized data of a randomly selected engine in FD002 are visualized in Fig.~\ref{normalization}. 

\subsection{Baselines}
\label{sec4.3}
Three baseline methods are considered. 

The first baseline is the functional data analysis-Based Empirical Bayesian method (`BayesFDA') \cite{zhou2011degradation}. The authors did not publish their code. We use the R package `fdapace'\footnote{https://CRAN.R-project.org/package=fdapace}, a well-developed functional data analytical package, to implement their method, including fitting the prior distribution and using the formula given in their theorem to calculate the posterior distribution. The expected curve of the posterior is treated as the forecasted health indicator curve. The major limitation of this method is that the validity of the posterior distribution calculation, i.e., the accuracy of the forecasting, heavily relies on whether the requirement in the theorem is satisfied. 



Another baseline formulates the health indicator forecasting problem from the regression perspective. The key idea is to use the first data source in Fig.~\ref{Fig:flow_chart} to train a model that outputs the next $V$ health indicator values using the past $U$ health condition measurements as the input. In our implementation, we set $U=30$ and $V=1$, and use the linear regression (`RG-Linear') and the long short-term memory (LSTM). Then the trained model together with the sliding window technique is applied to the new equipment initial health measurements to extrapolate the entire health indicator curve. It is well known that deep learning models typically require a large number of training samples to effectively learn complex mappings from noisy data. As shown in the experiments, given the limited amount of data, LSTM cannot learn a valid model regardless of the architecture of LSTM and the value of the hyperparameters. As LSTM does not work for the considered signals, we do not include it in the performance comparison.

To show the advantage of the generative modeling step in the proposed method, we consider a nearest neighbor based method (`NN'). In this baseline, we directly match each of the testing engines with the training engines and forecast the health curve by the one with the highest matching level in the training set. To eliminate the random noises, we also consider data smoothing for the achieved forecasts (`NN-S').

\subsection{Results}
\label{sec4.4}
In this experiment, we select four degradation signals (i.e., sensor $\#2$, $\#7$, $\#13$, $\#21$) and treat each of them as the health indicator. They are labeled as `Signal 1', `Signal 2', `Signal 3' and `Signal 4' in the following discussions. 

\textit{1) Evaluation metrics:} For a selected health indicator, suppose that the actual health indicator measurements over the lifespan are represented by $\{\tilde{\mathbf{S}}_{test, i}=[\tilde{S}_{test, i}(t_{i, 1}),...,\tilde{S}_{test, i}(t_{i, m_i})]^T\}_{i=1}^{N_{test}}$. For the $i$-th equipment, suppose that the first $m_i^*$ are given for curve forecasting models. The accuracy metric for health indicator forecasting result $\{\hat{\tilde{S}}_{test, i}(t)\}_{i=1}^{N_{test}}$ is 
\begin{equation}\label{res1}
\text{RMSE}_{ext} = \sqrt{\frac{\sum_{i=1}^{N_{test}}\sum_{j=m_i^*}^{m_{i}}(\tilde{S}_{test, i}(t_{i, j})-\hat{\tilde{S}}_{test, i}(t_{i, j}))^2}{\sum_{i=1}^{N_{test}}(m_{i}-m_i^*+1)}}.
\end{equation}

According to Eq.~(\ref{RUL2}) and (\ref{RUL3}) in Section \ref{sec3}, for the $i$-th equipment in testing, we can calculate the actual remaining time to a soft-failure $\text{RUL}_{i, m^*_i}$ and the predicted value $\hat{\text{RUL}}_{i, m^*_i}$ at $m^*_i$ corresponding to predetermined cutoff value $\theta$. The accuracy is evaluated by the root mean square error
\begin{equation}\label{res2}
\text{RMSE}_{rul} = \frac{1}{N_{test}}\sum_{i=1}^{N_{test}}(\text{RUL}_{i, m^*_i} - \hat{\text{RUL}}_{i, m^*_i})^2.
\end{equation}

For a given metric, the improvement of our proposed method over the best baseline is calculated by 
\begin{equation}
    \text{IMP}=1-\frac{\text{Metric of `proposal'}}{\text{Metric of the best baseline}}.\label{IMP1}
\end{equation}

\begin{figure}[htbp]
	\centering
	\begin{subfigure}[t]{1.65in}
		\centering
        \caption{Performance under different number of initial measurements.}\label{plot1}
        \vspace{-0.08in}
		\includegraphics[width=40mm]{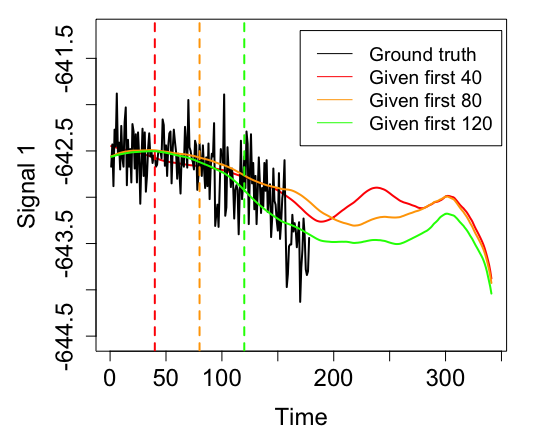}		
	\end{subfigure}
        \quad
	\begin{subfigure}[t]{1.65in}
		\centering
        \caption{Forecasting accuracy with different number of scenarios.}\label{plot2}
        \vspace{-0.08in}
		\includegraphics[width=40mm, height=30mm]{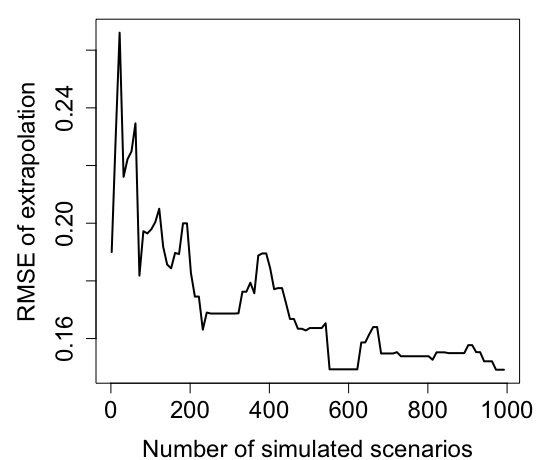}		
	\end{subfigure}
	\vspace{-0.05in}
	\caption{Investigation on the performance of our proposal.}\label{plot12}
\end{figure}

\textit{2) Investigation on our proposal:} Before comparing with the baselines, we investigate the performance of our proposal under various circumstances. First, as illustrated by Fig.~\ref{plot1}, the forecasted health indicator curve of our proposed method is capable of capturing the overall trend regardless of the amount of initial data given, while the accuracy increases when more initial data are supplied, which matches with our intuition.

To implement our proposed method, we need to determine how many random scenarios should be generated. As shown by Fig.~\ref{plot2}, the accuracy of the curve forecasting first decreases then stays relatively stable with the increase of the number of scenarios. This is because the generated scenarios are diverse enough to guarantee that a good matching with the new health curve exists after a certain number of simulations. In this experiment, we set the number of scenarios as $W=1000$.   

\textit{3) Performance comparison:} Results of the three baselines (`BayesFDA', `RG-Linear', `NN', and `NN-S') and our proposed `Generative + scenario matching' method (`Proposed') are summarized in Tab.~\ref{tab1} and \ref{tab2}. The major observations are summarized as follows:
\begin{itemize}[leftmargin=*]
\item For all circumstances, our proposed method significantly outperforms the other approaches, with an averaged $24\%$ improvement in health indicator curve forecasting and an averaged $20\%$ improvement in RUL estimation over the best baseline.
\item The nearest neighbor-based methods have the second-best performance, indicating that the training sets provide a relatively rich level of coverage for possible health indicator curve variations. This is reasonable as C-MAPSS data is synthetic data from a simulator built by NASA.
\item The functional data analysis based Empirical Bayesian method performs worse than our proposed method and the naive scenario matching approach. Our explanation is that the achieved forecasts are biased as the data does not satisfy the requirement discussed in Section \ref{sec1}.   
\item The linear regression-based method has the worst performance. Unlike the other methods that model the whole health indicator curves, it inappropriately cuts the health indicator curves into multiple windows and assumes that the extracted training data from different windows are independent even though they come from the different time periods of the same equipment. For a certain proportion of engines in the testing set, we cannot obtain a valid RUL estimation as the forecasted health indicators are all higher than the pre-defined threshold. The RUL estimation performance is not included in Tab.~\ref{tab2}. 
\end{itemize}



\begin{table}[htbp]
\caption{RMSE comparison on health indicator curve forecasting and improvement (`IMP') over the best baseline.}
\begin{center}
\vspace{-0.1in}
\scalebox{0.95}{
\begin{tabular}{cccccc}
\hline
\hline
\textbf{Data}& \textbf{Model}& \textbf{Signal 1}&   \textbf{Signal 2}&  \textbf{Signal 3}& \textbf{Signal 4}\\
\hline
FD001 & BayesFDA& 0.454 & 0.906 & 0.056  & 0.089  \\
      & RG-Linear & 0.495 & 1.024 & 0.061 &  0.134\\
      & NN & 0.204 & 0.441 & 0.042 &  0.046 \\
      & NN-S & 0.203 & 0.444 & 0.041 &  0.046 \\
      & Proposed & 0.153 & 0.250& 0.027 &   0.032 \\
\cline{2-6}
      & IMP& $25.00\%$ & $43.31\%$ & $35.71\%$  & $30.43\%$  \\
\hline
FD002 & BayesFDA& 0.470 & 0.463 & 0.235  & 0.065  \\
      & RG-Linear & 0.483 & 0.501 & 0.242 &  0.076 \\
      & NN & 0.200 & 0.245 & 0.141 &  0.031 \\
      & NN-S & 0.201 & 0.245 & 0.142 &  0.031 \\
      & Proposed & 0.170 & 0.204 & 0.132 &   0.025 \\
\cline{2-6}
      & IMP& $15.42\%$ & $16.73\%$ & $7.04\%$  & $19.35\%$  \\
\hline
\hline
\end{tabular}
}
\label{tab1}
\end{center}
\end{table}

\begin{table}[htbp]
\caption{RMSE comparison on RUL estimation and improvement (`IMP') over the best baseline.}
\begin{center}
\vspace{-0.1in}
\scalebox{0.925}{
\begin{tabular}{cccccc}
\hline
\hline
\textbf{Data}& \textbf{Model}& \textbf{Signal 1}&   \textbf{Signal 2}&  \textbf{Signal 3}& \textbf{Signal 4}\\
\hline
FD001 & BayesFDA& 161.91 & 163.65 & 139.23  & 120.34  \\
      & NN & 112.47 & 92.70 & 112.94 &  112.16 \\
      & NN-S & 43.13 & 38.25 & 64.12 &  42.97 \\
      & Proposed & 33.55 & 30.56 & 55.26 &   27.45 \\
\cline{2-6}
      & IMP& $22.21\%$ & $20.1\%$ & $13.82\%$  & $36.12\%$  \\
      &  & ($\theta$=643) & ($\theta$=552.8) & ($\theta$=2388.125)  & ($\theta$=23.28)  \\
\hline
FD002 & BayesFDA& 167.72 & 148.56 & 162.04  & 144.87  \\
      & NN & 149.23 & 135.87 & 153.23 &  150.92 \\
      & NN-S & 49.39 & 51.27 & 50.98 &  48.50 \\
      & Proposed & 46.52 & 47.57& 41.54 &   42.70 \\
\cline{2-6}
      & IMP& $6.18\%$ & $32.14\%$ & $18.52\%$  & $11.96\%$  \\
      &  & ($\theta$=0.125) & ($\theta$=0.125) & ($\theta$=0.1)  & ($\theta$=0.025)  \\      
\hline
\hline
\end{tabular}
}
\label{tab2}
\vspace{-0.1in}
\end{center}
\end{table}

\begin{figure}[htbp]
	\centering
	\begin{subfigure}[t]{1.65in}
		\centering
        \caption{Selected engine $\# 1$.}\label{plot3}
        \vspace{-0.08in}
		\includegraphics[width=35mm]{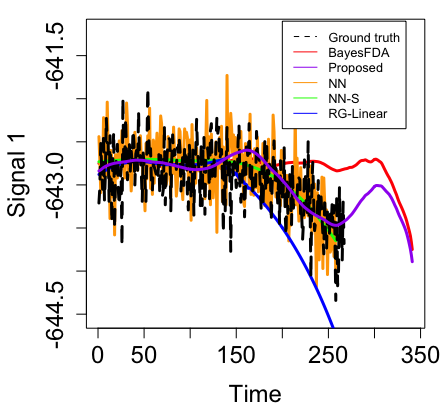}		
	\end{subfigure}
        \quad
	\begin{subfigure}[t]{1.65in}
		\centering
        \caption{Selected engine $\# 2$.}\label{plot4}
        \vspace{-0.08in}
		\includegraphics[width=35mm]{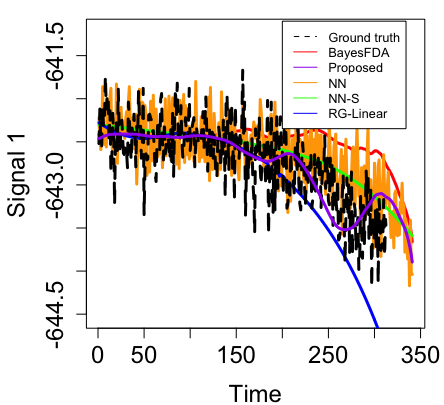}		
	\end{subfigure}
	\vspace{-0.05in}
	\caption{Forecasts for two randomly selected engines.}\label{plot34}
\end{figure}

\section{Conclusions and Discussions}
\label{sec5}
In data-driven prognostic, there is a growing demand to get accurate estimates for health indicator curves over an operating equipment lifespan. We proposed a new perspective to address the health indicator curve forecasting challenge. The proposed method is more general than the prior art, i.e., capable of achieving accurate forecasting for more diverse forms of health indicator data, due to the non-parametric techniques in the `prior' distribution fitting step and the proposed model-free `posterior' updating procedure. We propose a point estimator for soft-failures given the forecasted health indicator. Our experimental results on the well-known benchmark data set called NASA C-MAPSS data demonstrated that our proposed approach significantly outperforms alternative data-driven methods in terms of both health indicator forecasting and remaining useful life estimation.




\balance
\bibliographystyle{IEEEtran}
\bibliography{fanova}

\end{document}